\documentclass[11pt]{article}
\usepackage{eamt23}
\usepackage{times}
\usepackage{url}
\usepackage{latexsym}
\usepackage[small,bf]{caption} % added MLF 20171211
\usepackage{booktabs}

\title{Design of an Open-Source Architecture for Neural Machine Translation}

\author{Séamus Lankford\\
 ADAPT Centre, \\ MTU, Cork, Ireland\\
  {\tt seamus.lankford@mtu.ie}  \And
Haithem Afli\\
 ADAPT Centre, \\ MTU, Cork, Ireland\\
\hspace{1cm}  {\tt haithem.afli@mtu.ie}
   \And
Andy Way\\
 ADAPT Centre, \\ DCU, Dublin, Ireland\\
{\tt andy.way@dcu.ie}}

\date{}

\begin{document}
\maketitle
\begin{abstract}
adaptNMT is an open-source application that offers a streamlined approach to the development and deployment of Recurrent Neural Networks and Transformer models. This application is built upon the widely-adopted OpenNMT ecosystem, and is particularly useful for new entrants to the field, as it simplifies the setup of the development environment and creation of train, validation, and test splits. The application offers a graphing feature that illustrates the progress of model training, and employs SentencePiece for creating subword segmentation models. Furthermore, the application provides an intuitive user interface that facilitates hyperparameter customization. Notably, a single-click model development approach has been implemented, and models developed by adaptNMT can be evaluated using a range of metrics. To encourage eco-friendly research, adaptNMT incorporates a green report that flags the power consumption and kgCO\textsubscript2 emissions generated during model development. The application is freely available.\footnote{\url{http://github.com/adaptNMT}}
\end{abstract}

\section{Credits}
This research is supported by Science Foundation Ireland through the ADAPT Centre (Grant 13/RC/2106) (www.adaptcentre.ie) at Dublin City University. This research was also funded by the Munster Technological University.

\section{Introduction}\label{sec1}

Explainable Artificial Intelligence (XAI) \cite{arrieta2020explainable} aims to ensure that the outcomes of AI solutions are easily comprehensible to humans. In light of this goal, adaptNMT has been developed to provide users with a form of Explainable Neural Machine Translation (XNMT). The typical NMT process comprises several independent stages, including setting up the environment, preparing the dataset, training subword models, parameterizing and training the main models, evaluating and deploying them. By adopting a modular approach, this framework has established an effective NMT model development process that caters to both technical and non-technical practitioners in the field. To address the environmental impact of building and running large AI models \cite{henderson2020towards,jooste-etal-2022-knowledge}, we have also produced a ``green report'' that calculates carbon emissions. While primarily intended as an information aid, this report will hopefully encourage the development of reusable and sustainable models.

This research endeavors to create models and applications that address the challenges of language technology, which will be particularly beneficial for those new to the field of Machine Translation (MT) and those seeking to learn more about NMT.

The application is built on OpenNMT\footnote{\url{https://opennmt.net}}~\cite{klein2017opennmt} and thus inherits all of its features. Unlike many NMT toolkits, a command line interface (CLI) is not used, and the interface is designed and fully implemented in Google Colab.\footnote{\url{colab.research.google.com}} For both educational and research purposes, a cloud-hosted solution like Colab is often more user-friendly. Additionally, the training of models can be monitored and controlled via a Google Colab mobile app, which is useful for long-run builds. The adaptNMT framework also includes GUI controls that allow for the customization of all crucial parameters needed for NMT model training.

The application can be run in local mode to utilize existing infrastructure or hosted mode for rapid infrastructure scaling. A deploy function is also included to allow for the immediate deployment of trained models.

This paper begins by presenting background information on NMT and NMT tools in Section \ref{related}, followed by a detailed description of the adaptNMT architecture and its key features in Section \ref{arch}. The system is discussed in Section \ref{disc} before concluding with a discussion of future work in Section \ref{concl}. A more in-depth system description, coupled with an empirical evaluation
of models developed using the application, is outlined in a separate paper \cite{lankfordlrev}. 

\section{Related Work}\label{related}

\subsection{NMT}\label{subsec2}
In addition to the ongoing research dedicated to developing state-of-the-art (SOTA) NMT models, comprehensive descriptions of this technology are readily available in the literature, making it accessible to individuals who are new to the field or have limited technical expertise \cite{wayBlooms}.

NMT has benefitted from the availability of large parallel corpora, leading to the development of high-performing MT models. The field of MT has experienced significant advancements through the application of NMT, particularly after the introduction of the Transformer \cite{10.5555/3295222.3295349} architecture, which has resulted in SOTA performance across multiple language pairs \cite{bojar-etal-2017-findings,bojar-etal-2018-findings,lankford2021transformers,lankford-etal-2021-machine,lankford2022lrec,lankford2022human}.

\subsection{NMT Tools}\label{tools}

In essence, adaptNMT is an IPython wrapper built on OpenNMT, enabling it to benefit from OpenNMT's extensive feature set and continuous code maintenance. However, adaptNMT takes abstraction to a higher level than OpenNMT, with greater focus on usability, particularly for newcomers. As a result, adaptNMT facilitates easy and fast deployment, offering features such as more pre-processing, as well as GUI control over model creation. Moreover, it incorporates green features in line with current research efforts towards smaller models with reduced carbon footprints, making it suitable for educational and research environments alike.

Other commonly used frameworks for developing NMT systems include FAIRSEQ\footnote{\url{https://github.com/facebookresearch/fairseq}}~\cite{ott2019fairseq}, an open-source sequence modelling toolkit based on PyTorch that allows for training models for translation, summarization, and language modelling. Marian\footnote{\url{https://marian-nmt.github.io}}~\cite{junczys2018marian}, on the other hand, is an NMT framework based on dynamic computation graphs and developed using C++. OpenNMT is an open-source NMT framework that has been widely adopted in the research community and covers the entire MT workflow from data preparation to live inference.

\section{Architecture of adaptNMT}\label{arch}

\begin{figure*}[ht]
    \centering
    \includegraphics[width=.95\linewidth]{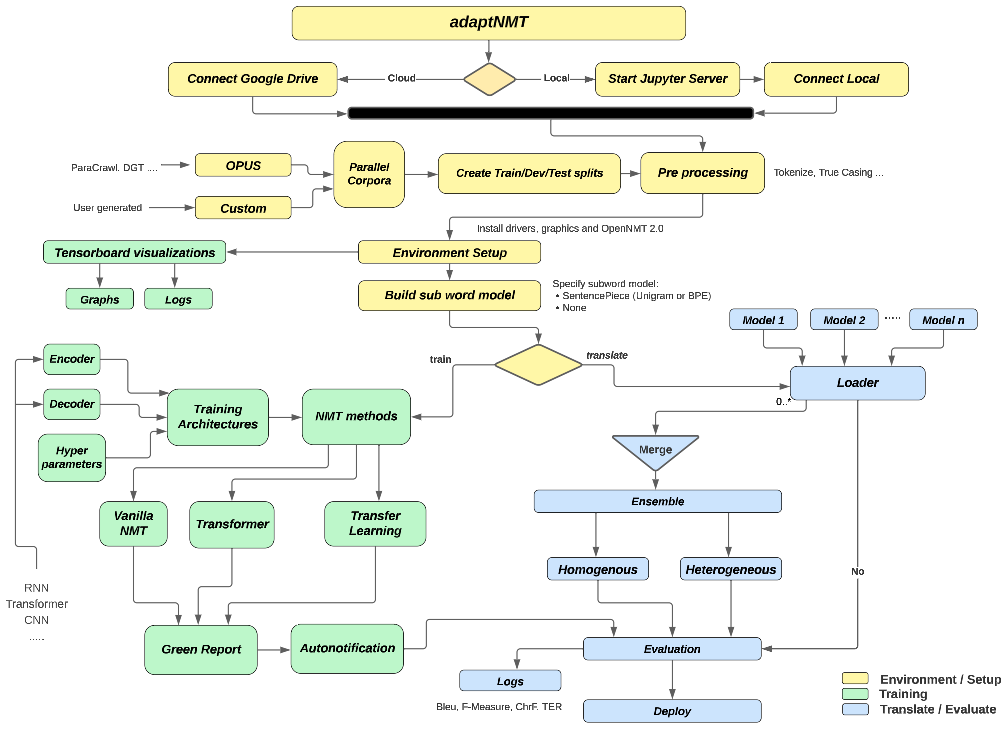}
    \caption{\footnotesize {Proposed architecture for adaptNMT: a language-agnostic NMT development environment. The system is designed to run either in the cloud or using local infrastructure. Models are trained using parallel corpora. Visualization and extensive logging enable real-time monitoring. Models are developed using vanilla RNN-based NMT, Transformer-based approaches or transfer learning using a fine-tuning approach. Translation and evaluation can be carried out using either single models or ensembles.}}
    \label{fig:approach}
\end{figure*}

After providing a general overview of NMT and NMT development systems, we introduce the adaptNMT tool, which enables users to configure the components of the NMT development process. The platform's system architecture is depicted in Figure \ref{fig:approach}. The tool is built as an IPython notebook and leverages the Pytorch implementation of OpenNMT for training models. Additionally, SentencePiece is used to train subword models. Using a Jupyter notebook facilitates sharing the application with other members of the MT community, and the application's setup is simplified since all necessary packages are downloaded dynamically as the application runs.

The system has two deployment options: running it locally or as a Colab instance via Google Cloud. In order to build translation models, the system requires parallel text corpora for both the source and target languages. A Tensorboard visualization allows for real-time monitoring of the model training process. At runtime, users can select to use the system for either model building or translation services, or both. Additionally, as depicted in Figure \ref{fig:approach}, the system enables the generation of an ensemble output during translation. Finally, trained models can be easily deployed to a pre-configured location.

\subsection{adaptNMT}\label{aNMT} 
The application may be run as an IPython Jupyter notebook or as a Google Colab application. Given the ease of integrating large Google drive storage into Colab, the application has been used exclusively as a Google Colab application for our own experiments. 

\subsubsection{Initialization and logging}

Initialization enables connection to Google Drive to run experiments, automatic installation of Python, OpenNMT,\footnote{\url{https://opennmt.net}} SentencePiece,\footnote{\url{ https://github.com/google/sentencepiece}} Pytorch and other applications. The visualization section enables real-time graphing of  model development. All log files are stored and can be viewed to inspect training convergence, the model’s training and validation accuracy and changes in learning rates.

\subsubsection{Modes of operation}
There are two modes of operation: local or cloud. In local mode, the application is run so that models are built using the user's local GPU resources. The option to use cloud mode enables users to develop models using Google's GPU clusters. For shorter training times, the unpaid Colab option is adequate. However, for a small monthly subscription, the Google Colab Pro option is worthwhile since users have access to improved GPU and compute resources. Furthermore, using Google Cloud may be considered as the ``green option'' since its platform uses 100\% renewables \cite{lacoste2019quantifying}. 

\subsubsection{Customization of models}
The system has been developed to allow users to select variations to the underlying model architecture. A vanilla RNN or Transformer approach may be selected to develop the NMT model. The customization mode enables users to specify the exact parameters required for the chosen approach. One of the features, AutoBuild, enables a user to build an NMT model in three simple steps: (i) upload source and target files, (ii) select  RNN or Transformer, and (iii) click AutoBuild.

\subsubsection{Use of subword segmentation}
In the NMT development process, users can specify the type of optimizer for learning and choose from different subword models. The subword model functionality allows for the selection of a subword model type and the choice of vocabulary size, currently offering either a SentencePiece unigram or a SentencePiece BPE model.

A user may upload a dataset which includes the train, validation and test splits for both source and target languages. In cases where a user has not already created the required splits for model training, single source and target files may be uploaded. Automated splitting of the uploaded dataset into train, validation, and test files is then performed based on the user's chosen split ratio. 

Given that building NMT models typically demands long training times, an automatic notification feature is incorporated that informs the user by email when model training has been completed.

\subsubsection{Translation and evaluation}

The application supports not only the training of models but also the translation and evaluation of model performance. For translation using pre-built models, users can specify the model name as a hyperparameter which is subsequently used to translate and evaluate the test files. The option for creating an ensemble output is also available, with users simply naming the models to be used in generating the ensemble output.

Once the system has been built, the user can select the model to be used for translating the test set. While human evaluation is often considered the most insightful approach for evaluating translation quality, it can be limited by factors such as availability, cost, and subjectivity. Thus, automatic evaluation metrics are frequently employed, particularly by developers monitoring incremental progress of systems. A further discussion on the advantages and disadvantages of human and automatic evaluation is available in the literature \cite{Way2018}.

Several automatic evaluation metrics provided by SacreBleu\footnote{\url{https://github.com/mjpost/sacrebleu}} \cite{post2018call} are used: BLEU \cite{papineni2002bleu}, TER \cite{snover2006study} and ChrF \cite{popovic2015chrf}. Translation quality can also be evaluated using Meteor~\cite{denkowski2014meteor} and F1 score~\cite{melamed-etal-2003-precision}. Note that BLEU, ChrF, Meteor and F1 are precision-based metrics, so higher scores are better, whereas TER is an error-based metric and lower scores indicate better translation quality. Evaluation options available include standard (truecase) and lowercase BLEU scores, a sentence-level BLEU score option, ChrF1 and ChrF3.  

There are three levels of logging: model development logs for graphing, training console output and experimental results. A references section outlines resources which are relevant to developing, using and understanding adaptNMT. Validation during training is currently conducted using model accuracy and perplexity (PPL). 

\subsection{Infrastructure}
Rapid prototype development is possible through a Google Colab Pro subscription using NVIDIA Tesla P100 PCIe 16GB graphic cards and up to 27GB of memory when available. 

\section{Discussion}\label{disc}

Numerous tools have been developed to assess the carbon footprint of NLP \cite{bannour2021evaluating}. The notion of sustainable NLP has also gained momentum as an independent research track, with high-profile conferences such as the \textit{EACL 2021 Green and Sustainable NLP} track dedicating resources to this area.\footnote{\url{https://2021.eacl.org/news/green-and-sustainable-nlp}} 

Given these developments, we have incorporated a "green report" into adaptNMT that logs the kgCO\textsubscript2 generated during model development. This aligns with the industry's increasing focus on quantifying the environmental impact of NLP. In fact, it has been demonstrated that high-performing MT systems can be developed with much lower carbon footprints, leading to significant energy cost savings for a real translation company \cite{info13020088}.

The risks associated with relying on Large Language Models (LLMs) have been well-documented in the literature. The discussion surrounding these models emphasizes not only their environmental impact but also the inherent biases and dangers they pose for low-resource languages \cite{bender2021dangers}. It is important to note that smaller, in-domain datasets can yield high-performing NMT models, and the adaptNMT framework makes this approach easily accessible and understandable.

\section{Conclusion and Future Work}\label{concl}

We have introduced adaptNMT, an application that manages the entire NMT model development, evaluation, and deployment workflow.

As for future work, our development efforts will be directed towards incorporating new transfer learning methods and improving our ability to track environmental costs. We will integrate modern zero-shot and few-shot approaches, as seen in the GPT3 \cite{brown2020language} and Facebook LASER \cite{artetxe2019massively} frameworks. While the existing adaptNMT application is focused on customizing NMT models, we will also develop a separate application, adaptLLM \cite{lankford2023adaptLLM,lankford-mtsummit-2023-llms}, for fine-tuning LLMs. In particular, adaptLLM will cater for low-resource language pairs such as NLLB \cite{costa2022no}.

The green report integrated into the application represents our first implementation of a sustainable NLP feature within adaptNMT. We plan to enhance this feature by improving the user interface and providing recommendations on how to develop greener models. As an open-source project, we invite the community to contribute new ideas and improvements to the development of this feature.

\bibliography{eamt23}
\bibliographystyle{eamt23}
\end{document}